# A Novel Transformer based Semantic Segmentation Scheme for Fine-Resolution Remote Sensing Images

Libo Wang, Rui Li, Chenxi Duan, Ce Zhang, Xiaoliang Meng and Shenghui Fang

*Abstract*—The Fully Convolutional Network (FCN) with an encoder-decoder architecture has been the standard paradigm for semantic segmentation. The encoder-decoder architecture utilizes an encoder to capture multi-level feature maps, which are incorporated into the final prediction by a decoder. As the context is crucial for precise segmentation, tremendous effort has been made to extract such information in an intelligent fashion, including employing dilated/atrous convolutions or inserting attention modules. However, these endeavours are all based on the FCN architecture with ResNet or other backbones, which cannot fully exploit the context from the theoretical concept. By contrast, we introduce the Swin Transformer as the backbone to extract the context information and design a novel decoder of densely connected feature aggregation module (DCFAM) to restore the resolution and produce the segmentation map. The experimental results on two remotely sensed semantic segmentation datasets demonstrate the effectiveness of the proposed scheme. Code is available at https://github.com/WangLibo1995/GeoSeg.

*Index Terms*—semantic segmentation, fine-resolution remote sensing images, transformer

## I. INTRODUCTION

As an effective method to extract features automatically and hierarchically from images, the convolutional neural network (CNN) has become the common framework for tasks related to computer vision (CV) [4]. For semantic segmentation, the Fully Convolutional Network (FCN) [7] is the first proven and effective end-to-end CNN structure. Specifically, there are two symmetric paths in the FCN and its variants: a contracting path, i.e., the encoder, for extracting features, and an expanding path, i.e., the decoder, for exacting positions [10]. The contracting path, by definition, gradually downsamples the resolution of feature maps to reduce the computational consumption, while the expanding path can learn more semantic meaning via a progressively increasing receptive field. Benefit from its translation equivariance and locality, the FCN enhances the segmentation performance significantly and influences the entire field. Specifically, the translation equivariance underpins the generalization capability of the model to unseen data, while the locality reduces the complexity of the model by sharing parameters.

The outcome of FCN, although encouraging, appears to be coarse due to the over-simplified design of the decoder. Subsequently, more elaborate encoder-decoder structures were proposed [17], thus increasing the accuracy further. However, the long-range dependency is limited by the locality property of FCN-based methods, which is critical for segmentation in unconstrained scene images. There are two types of methods to address the issue, either modifying the convolution operation or utilizing the attention mechanism. The former aiming to enlarge the receptive fields using large kernel sizes [18], dilated convolutions [19], or feature pyramids [2, 20], whereas the latter focuses on integrating attention mechanisms with the FCN architecture to capture long-range dependencies of the feature maps [5, 21]. Nevertheless, both methods fail to liberate the network from the dependence of the FCN structure. More recently, several inspiring advances [22, 23] attempt to avoid convolution operations completely by employing attention-alone models, thereby achieving feature maps with long-range dependencies effectively.

For natural language processing (NLP), the dominant architecture is the Transformer [24], which adopts the multi-head attention to model long-range dependencies for sequence modelling and transduction tasks. The tremendous breakthrough in the natural language domain inspires researchers to explore the potential and feasibility of Transformer in the computer vision field. Obviously, the successful application of Transformer will become the first and foremost step to integrate computer vision and NLP, thereby providing a universal and uniform artificial intelligence (AI) scheme.

The pioneering work of Swin Transformer [22] presents a hierarchical feature representation scheme that demonstrates impressive performances with linear computational complexity. In this Letter, we *first* introduce the Swin Transformer for semantic segmentation of fine-resolution remote sensing images. Most importantly, we propose a densely connected

This work was funded by National Natural Science Foundation of China (NSFC) under grant number 41971352. *(Corresponding author: Shenghui Fang.)*

L. Wang, R. Li, X. Meng and S. Fang are with School of Remote Sensing and Information Engineering, Wuhan University, Wuhan 430079, China (e-mail: wanglibo@whu.edu.cn; lironui@whu.edu.cn; xmeng@whu.edu.cn; shfang@whu.edu.cn).

C. Duan is with the State Key Laboratory of Information Engineering in Surveying, Mapping, and Remote Sensing, Wuhan University, Wuhan 430079, China; chenxiduan@whu.edu.cn (e-mail: chenxiduan@whu.edu.cn).

C. Zhang is with Lancaster Environment Centre, Lancaster University, Lancaster LA1 4YQ, United Kingdom; UK Centre for Ecology & Hydrology, Library Avenue, Lancaster, LA1 4AP, United Kingdom (e-mail: c.zhang@lancaster.ac.uk).

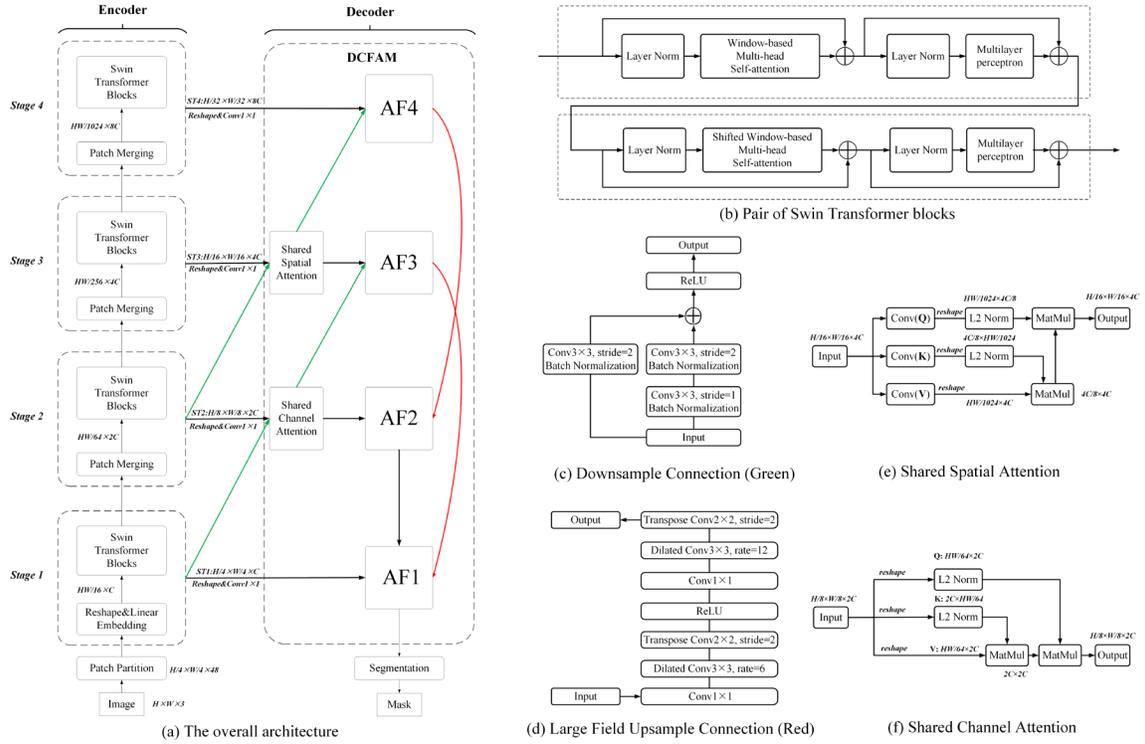

**Fig. 1** (a) The overall architecture of DC-Swin, (b) Pair of Swin Transformer Blocks, (c) Downsample Connection, (d) Large Field Upsample Connection, (e) Shared Spatial Attention, and (f) Shared Channel Attention. The values of H and W are both 1024. Please enlarge the PDF to >=200% to get a better view.

feature aggregation module (DCFAM) to extract multi-scale relation-enhanced semantic features for precise segmentation. Combining Swin Transformer and DCFAM, a novel semantic segmentation scheme of Densely Connected Swin Transformer (DC-Swin) is established.

## II. METHODOLOGY

The overall architecture of our DC-Swin is constructed based on the encoder-decoder structure, where the Swin Transformer is introduced as the encoder while the proposed DCFAM is selected as the decoder.

### A. Swin Transformer

As shown in Fig.1 (a), the Swin Transformer backbone [22] first utilizes a patch partition module to split the input RGB image into non-overlapping patches as "tokens". The feature of each patch is set as a concatenation of the raw pixel RGB values. Subsequently, this raw-valued feature is fed into the multistage feature transformation. In stage 1, a linear embedding layer is deployed to project features to an arbitrary dimension $C$. Thereafter, pairs of Swin Transformer blocks (Fig.1 (b)), which can maintain the number of tokens (e.g., $HW/16$), are adopted to extract semantic features. In the remaining stages, the number of tokens is gradually reduced by patch merging layers along with the increasing depth of the network to produce a hierarchical representation. The outputs of the four stages are processed by a standard $1 \times 1$ convolution to generate four hierarchical Swin Transformer features (**$ST_1$, $ST_2$, $ST_3$, and $ST_4$**).

By choosing diverse hyper-parameters, i.e., the dimensions $C$ and the number of Swin Transformer blocks in each stage, four Swin Transformer backbones with different complexities can be obtained:

- Swin-T: $C = 96$, block numbers = {2, 2, 6, 2}
- Swin-S: $C = 96$, block numbers = {2, 2, 18, 2}
- Swin-B: $C = 128$, block numbers = {2, 2, 18, 2}
- Swin-L: $C = 192$, block numbers = {2, 2, 18, 2}

In this letter, to balance the efficiency and effectiveness, we choose Swin-S pre-trained on the ImageNet as the backbone of the encoder, with the number of parameters (50M) comparable to ResNet-101 (45M).

### B. Densely Connected Feature Aggregation Module

Multi-scale and confusing geospatial objects appear frequently in fine-resolution remote sensing images, which seriously affects the quality of segmentation. To handle this issue, we propose a novel DCFAM method for feature representation. To be specific, we design a Shared Spatial Attention (SSA) and a Shared Channel Attention (SCA) to enhance the spatial-wise and channel-wise relationship of the semantic features based on our previous work of linear attention mechanism [25]. Besides, multi-level features are further integrated using the Downsample Connection and the Large-field Upsample Connection for improving multi-scale representation. As shown in Fig.1, the DCFAM connects the four hierarchical transformer features with cross-scale connections (i.e., Downsample Connection and Large Field Upsample Connection) and attention blocks (i.e., Shared Spatial Attention and Shared Channel Attention), generating four aggregation features (i.e., **$AF_1$, $AF_2$, $AF_3$, and $AF_4$**). Capitalising on the benefits provided by the DCFAM, the final segmentation feature **$AF_1$** is abundant in multi-scale information and relation-enhanced context.

*Downsample Connection:* The Downsample connection aims to connect the low-level and high-level transformer features for fusion, which can be defined as follow:

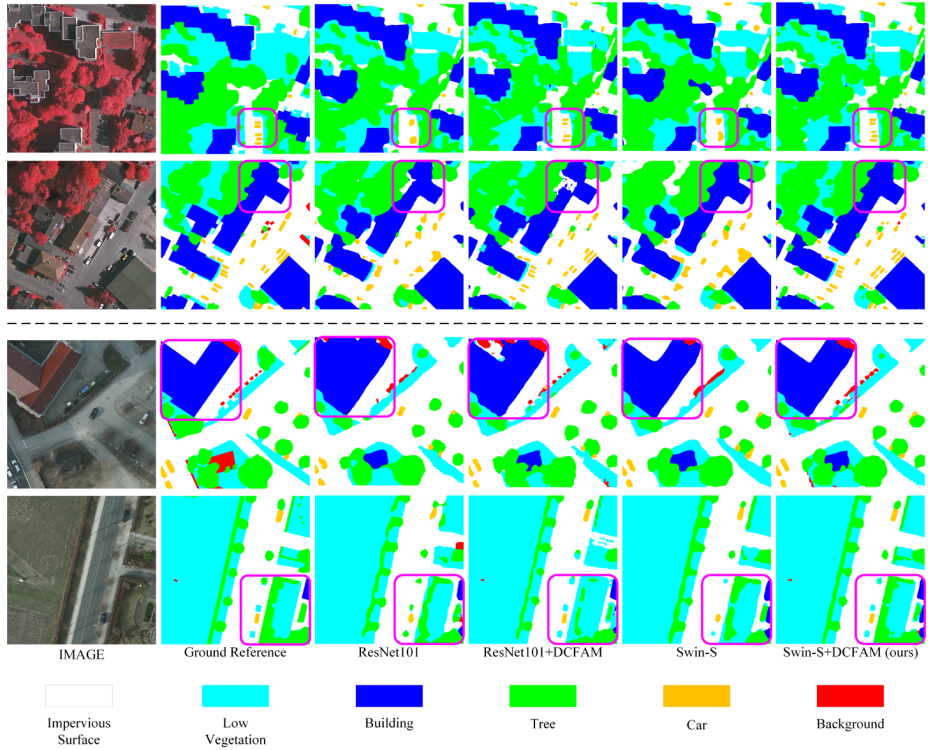

| Impervious Surface | Low Vegetation | Building | Tree | Car | Background |

**Fig. 2** Enlarged visualization of results on the Vaihingen dataset (Top) and Potsdam dataset (Bottom).

$$D_i^j(\mathbf{X}) = f_\sigma(f_\delta(\mathbf{X}) + f_\mu(f_\theta(\mathbf{X}))) \tag{1}$$

where $\mathbf{X}$ is the input vector. $f_\sigma$ is a ReLU activation function. $f_\delta$ and $f_\mu$ are a 3×3 convolution layer with a stride of 2, $f_\theta$ is a 3×3 convolution layer with a stride of 1, and each convolution layer involves a batch normalization operation. $i$ and $j$ denote the number of the input channels and output channels, respectively.

*Large field Upsample Connection:* To capture multi-scale context effectively, we embedded the dilated convolution into the Large filed Upsample Connection formulated as:

$$LU_m^n(\mathbf{X}) = f_\varphi^{12}(f_\sigma(f_\varphi^6(\mathbf{X}))) \tag{2}$$

where $f_\varphi^{12}$ is a composite function that contains a standard 1×1 convolution, a dilated convolution with a dilated rate of 12, and a standard transpose convolution. Similarly, $f_\varphi^6$ has a dilated rate of 6. $m$ and $n$ represent the number of the input channel and output channel, respectively.

*Shared Spatial Attention:* Based on the linear attention mechanism [25], we utilize the Shared Spatial Attention to model the long-range dependencies in the spatial dimension defined as:

$$SSA(\mathbf{X})$$

$$= \frac{\sum_n V(\mathbf{X})_{c,n} + \left(\frac{Q(\mathbf{X})}{\|Q(\mathbf{X})\|_2}\right)\left(\left(\frac{K(\mathbf{X})}{\|K(\mathbf{X})\|_2}\right)^T V(\mathbf{X})\right)}{N + \left(\frac{Q(\mathbf{X})}{\|Q(\mathbf{X})\|_2}\right)\sum_n \left(\frac{K(\mathbf{X})}{\|K(\mathbf{X})\|_2}\right)^T_{c,n}} \tag{3}$$

where $Q(\mathbf{X})$, $K(\mathbf{X})$, and $V(\mathbf{X})$ represent the convolutional operation to generate the *query* matrix $\mathbf{Q} \in \mathbb{R}^{N \times D_k}$, *key* matrix $\mathbf{K} \in \mathbb{R}^{N \times D_k}$, and *value* matrix $\mathbf{V} \in \mathbb{R}^{N \times D_v}$. $N$ is the number of pixels in the input feature maps. $c$ and $n$ indicate the channel dimension and the flattened spatial dimension.

*Shared Channel Attention:* Similarly, the Shared Channel Attention is designed to extract the long-range dependencies among the channel dimension:

$$SCA(\mathbf{X})$$

$$= \frac{\sum_c R(\mathbf{X})_{c,n} + \left(R(\mathbf{X})_{c,n}\left(\frac{R(\mathbf{X})}{\|R(\mathbf{X})\|_2}\right)^T\right)\frac{R(\mathbf{X})}{\|R(\mathbf{X})\|_2}}{N + \left(\frac{R(\mathbf{X})}{\|R(\mathbf{X})\|_2}\right)^T \sum_c \left(\frac{R(\mathbf{X})}{\|R(\mathbf{X})\|_2}\right)^T_{c,n}} \tag{4}$$

where $R(\mathbf{X})$ indicate the reshape operation to flatten the spatial dimension. The detailed information about our previous work on the linear attention mechanism can be referred to [25].

*Feature aggregation:* The four aggregation features (**AF₁**, **AF₂**, **AF₃**, and **AF₄**) can eventually be computed by the following equations:

$$\mathbf{AF_4} = \mathbf{ST_4} + D_{384}^{768}(SSA(D_{192}^{384}(\mathbf{ST_2}))) \tag{5}$$

$$\mathbf{AF_3} = SSA(\mathbf{ST_3}) + D_{192}^{384}(SCA(D_{96}^{192}(\mathbf{ST_1}))) \tag{6}$$

$$\mathbf{AF_2} = SCA(\mathbf{ST_2}) + LU_{768}^{192}(\mathbf{AF_4}) \tag{7}$$

$$\mathbf{AF_1} = \mathbf{ST_1} + U(\mathbf{AF_2}) + LU_{384}^{96}(\mathbf{AF_3}) \tag{8}$$

Here, $U$ is a bilinear interpolation upsample operation with a scale factor of 2.

## III. EXPERIMENTAL RESULTS

### A. Dataset

We test the effectiveness of the proposed scheme on the well-known ISPRS Vaihingen and Potsdam semantic labelling datasets. There are 33 tiles extracted from true orthophoto and the co-registered normalized DSMs in the Vaihingen dataset with an average size of $2494 \times 2064$ pixels. The Potsdam dataset contains 38 tiles and the size of each tile is $6000 \times 6000$. Following previous pieces of literature [3, 9, 11], in the Vaihingen dataset, we use the benchmark organizer defined 16 images for training and 17 for testing, while the setting in the Potsdam dataset is 24 tiles for training and 14 tiles for testing. The image tiles are cropped into $1024 \times 1024$ px patches as the input. We do not employ DSMs in our experiments to reduce computation.

TABLE I
THE EXPERIMENTAL RESULTS ON THE VAIHINGEN DATASET.

| Method | Backbone | Imp. surf. | Building | Low veg. | Tree | Car | Mean F1 | OA | mIoU |
|---|---|---|---|---|---|---|---|---|---|
| DeepLabV3+ [1] | ResNet101 | 92.38 | 95.17 | 84.29 | 89.52 | 86.47 | 89.57 | 90.56 | 81.47 |
| PSPNet [2] | ResNet101 | 92.79 | 95.46 | 84.51 | 89.94 | **88.61** | 90.26 | 90.85 | 82.58 |
| DANet [5] | ResNet101 | 91.63 | 95.02 | 83.25 | 88.87 | 87.16 | 89.19 | 90.44 | 81.32 |
| EaNet [8] | ResNet101 | 93.40 | **96.20** | _85.60_ | **90.50** | _88.30_ | **90.80** | _91.20_ | - |
| DDCM-Net [3] | ResNet50 | 92.70 | 95.30 | 83.30 | 89.40 | _88.30_ | 89.80 | 90.40 | - |
| CASIA2 [11] | ResNet101 | 93.20 | 96.00 | 84.70 | 89.90 | 86.70 | 90.10 | 91.10 | - |
| V-FuseNet [9] | FuseNet | 91.00 | 94.40 | 84.50 | 89.90 | 86.30 | 89.20 | 90.00 | - |
| DLR_9 [15] | - | 92.40 | 95.20 | 83.90 | 89.90 | 81.20 | 88.50 | 90.30 | - |
| BoTNet [14] | ResNet50 | 92.24 | 95.28 | 83.88 | 89.99 | 85.47 | 89.37 | 90.51 | 81.05 |
| ResT [16] | ResT-Base | 92.15 | 94.88 | 84.17 | 90.02 | 84.97 | 89.24 | 90.43 | 80.82 |
| Ours | Swin-S | **93.60** | _96.18_ | **85.75** | _90.36_ | 87.64 | **90.71** | **91.63** | **83.22** |

TABLE II
THE EXPERIMENTAL RESULTS ON THE POTSDAM DATASET.

| Method | Backbone | Imp. surf. | Building | Low veg. | Tree | Car | Mean F1 | OA | mIoU |
|---|---|---|---|---|---|---|---|---|---|
| DeepLabV3+ [1] | ResNet101 | 92.95 | 95.88 | 87.62 | 88.15 | 96.02 | 92.12 | 90.88 | 84.32 |
| PSPNet [2] | ResNet101 | 93.36 | 96.97 | 87.75 | 88.50 | 95.42 | 92.40 | 91.08 | 84.88 |
| DDCM-Net [3] | ResNet50 | 92.90 | 96.90 | 87.70 | _89.40_ | 94.90 | 92.30 | 90.80 | - |
| CCNet [6] | ResNet101 | 93.58 | 96.77 | 86.87 | 88.59 | _96.24_ | 92.41 | 91.47 | 85.65 |
| AMA_1 | - | 93.40 | 96.80 | 87.70 | 88.80 | 96.00 | _92.54_ | 91.20 | - |
| SWJ_2 | ResNet101 | **94.40** | _97.40_ | _87.80_ | 87.60 | 94.70 | 92.38 | _91.70_ | - |
| V-FuseNet [9] | FuseNet | 92.70 | 96.30 | 87.30 | 88.50 | 95.40 | 92.04 | 90.60 | - |
| DST_5 [12] | FCN | 92.50 | 96.40 | 86.70 | 88.00 | 94.70 | 91.66 | 90.30 | - |
| BoTNet [14] | ResNet50 | 93.13 | 96.37 | 87.31 | 88.01 | 95.79 | 92.12 | 90.76 | 85.62 |
| ResT [16] | ResT-Base | 92.74 | 96.08 | 87.48 | 88.55 | 94.76 | 91.92 | 90.57 | 85.23 |
| Ours | Swin-S | _94.19_ | **97.57** | **88.57** | **89.62** | **96.31** | **93.25** | **92.00** | **87.56** |

### B. Experimental Setting

All of the experiments are implemented with PyTorch on a single RTX 3090, and the optimizer is set as AdamW with a 0.0003 learning rate. The soft cross-entropy is used as the loss function. For each method, the overall accuracy (OA), mean Intersection over Union (mIoU), and F1-score (F1) are chosen as evaluation metrics.

$$OA = \frac{\sum_{k=1}^{N} TP_k}{\sum_{k=1}^{N} TP_k + FP_k + TN_k + FN_k}, \quad (9)$$

$$mIoU = \frac{1}{N} \sum_{k=1}^{N} \frac{TP_k}{TP_k + FP_k + FN_k}, \quad (10)$$

$$precision = \frac{1}{N} \sum_{k=1}^{N} \frac{TP_k}{TP_k + FP_k}, \quad (11)$$

$$recall = \frac{1}{N} \sum_{k=1}^{N} \frac{TP_k}{TP_k + FN_k}, \quad (12)$$

$$F1 = 2 \times \frac{precision \times recall}{precision + recall}, \quad (13)$$

where $TP_k$, $FP_k$, $TN_k$, and $FN_k$ indicate the true positive, false positive, true negative, and false negatives, respectively, for the specific object indexed as class $k$. OA is computed for all categories including the background.

### C. Semantic Segmentation Results and Analysis

*1) Performance Comparison:* The experimental results on the Vaihingen and Potsdam datasets among state-of-the-art methods are listed in Table I and Table II. The quantitive indices demonstrate the effectiveness of the proposed segmentation scheme. To be specific, our proposed DC-Swin achieves 90.71% in mean F1-score, 91.63% in OA, and 83.22% in mIoU for the Vaihingen dataset, with 93.25%, 92.00%, and 87.56% for the Potsdam dataset, outperforming the majority of ResNet-based methods with highly competitive accuracy. Benefit from the global context information modelled by the Swin-S and the DCFAM, the performance of our scheme not only outperforms recent contextual information aggregation methods designed initially for natural images, such as DeepLabV3+ and PSPNet, but also prevails over the lastest multi-scale feature aggregation models proposed for remote sensing images, such as EaNet and DDCM-Net, as well as the transformer networks BoTNet and ResT.

*2) Ablation Study:* As we not only propose a novel feature aggregation model but also introduce a brand-new backbone for segmentation, it is valuable to conduct the ablation study and investigate the contribution of each part upon accuracy. For the ablation study, we select ResNet-101 and Swin-S with the direct upsample operation as the baseline. ResNet101+DC and Swin-S+DC, which remove the SCA and SSA from DCFAM, are developed for the ablation study of dense connections. DCFAM-NS denotes the modified DCFAM that adopts the no-shared form structure. As shown in Table III, the substitution of the backbone from ResNet-101 to Swin-S yields a 3% increase in the Vaihingen dataset and a 4.05% increase in the Potsdam dataset for the mIoU index, showing the superiority of Swin-S. ResNet101+DC and Swin-S+DC improve the performance of the corresponding baseline method dramatically, indicating the effectiveness of dense connections. Meanwhile, deploying the shared attention modules in DCFAM further increases the accuracy, demonstrating the effectiveness of the SCA and SSA. Besides, the employment of DCFAM-NS obtains lower scores compared to the utilization of DCFAM, which demonstrates the advantage of our shared form structure. Benefiting from the long-range dependencies and shared multi-scale structure, Swin-S+DCFAM obtains the highest accuracy on the two datasets, whose performance can also be observed in Fig. 2.

TABLE III
ABLATION STUDY ON THE VAIHINGEN AND POTSDAM DATASETS.

| Dataset | Method | Mean F1 | OA | mIoU |
|---------|--------|---------|-----|------|
| Vaihingen | ResNet101 | 85.31 | 89.59 | 75.48 |
| | ResNet101+DC | 88.96 | 90.73 | 80.48 |
| | ResNet101+DCFAM-NS | 89.48 | 90.87 | 81.26 |
| | ResNet101+DCFAM | 90.22 | 91.04 | 82.43 |
| | Swin-S | 87.54 | 90.50 | 78.48 |
| | Swin-S+DC | 88.91 | 91.11 | 81.94 |
| | Swin-S+DCFAM-NS | 89.96 | 91.26 | 82.02 |
| | Swin-S+DCFAM | **90.71** | **91.63** | **83.22** |
| Potsdam | ResNet101 | 88.66 | 89.24 | 79.97 |
| | ResNet101+DC | 91.75 | 90.45 | 84.95 |
| | ResNet101+DCFAM-NS | 91.81 | 90.49 | 85.05 |
| | ResNet101+DCFAM | 92.28 | 90.81 | 85.87 |
| | Swin-S | 91.20 | 90.54 | 84.02 |
| | Swin-S+DC | 92.55 | 91.33 | 86.32 |
| | Swin-S+DCFAM-NS | 92.82 | 91.47 | 86.80 |
| | Swin-S+DCFAM | **93.25** | **92.00** | **87.56** |

## IV. CONCLUSION

In this Letter, for the first time, we introduce Transformer into semantic segmentation of fine-resolution remote sensing images. We develop a densely connected feature aggregation module to capture multi-scale relation-enhanced semantic features, thereby increasing the segmentation accuracy. Numerical experiments conducted on the ISPRS Vaihingen and Potsdam datasets demonstrate the effectiveness of our scheme in segmentation accuracy. We envisage this pioneering Letter could inspire researchers and practitioners in this field to explore the potential and feasibility of the Transformer more widely in the remote sensing and Earth observation domain.